\documentclass[11pt,a4paper]{article}
\usepackage{coling2020}
\usepackage{times}  
\usepackage{helvet}  
\usepackage{courier}  
\usepackage{url}  
\usepackage{graphicx}  

\usepackage[symbol]{footmisc}
\usepackage{natbib}
\usepackage{bm}
\usepackage{amsmath,amsfonts,amssymb,amsthm}
\usepackage{tikz}
\usetikzlibrary{arrows, shapes, decorations.pathreplacing, calc, positioning}
\usepackage{tkz-graph}
\usepackage{algorithm}
\usepackage{algpseudocode}
\usepackage{caption}
\usepackage{subcaption}
\usepackage{enumitem}
\usepackage{tabularx}
\usepackage{bbm}
\usepackage{tablefootnote}
\usepackage{tikz-dependency}
\colingfinalcopy 

\makeatother
\setlist{nolistsep}

\tikzset{
    ncbar angle/.initial=90,
    ncbar/.style={
        to path=(\tikztostart)
        -- ($(\tikztostart)!#1!\pgfkeysvalueof{/tikz/ncbar angle}:(\tikztotarget)$)
        -- ($(\tikztotarget)!($(\tikztostart)!#1!\pgfkeysvalueof{/tikz/ncbar angle}:(\tikztotarget)$)!\pgfkeysvalueof{/tikz/ncbar angle}:(\tikztostart)$)
        -- (\tikztotarget)
    },
    ncbar/.default=0.5cm,
}
\tikzset{round left paren/.style={ncbar=0.5cm,out=120,in=-120}}
\tikzset{round right paren/.style={ncbar=0.5cm,out=60,in=-60}}

\newtheorem{theorem}{Theorem}[section]

\newtheorem{lemma}[theorem]{Lemma}

\title{Semi-supervised Autoencoding Projective Dependency Parsing}

\author{Xiao Zhang\thanks{\quad This work was majorly done while the first author was pursuing a Ph.D. degree at Purdue University.} \\
  Search Science \& AI\\
  Amazon.com \\
  {\tt zhxao@amazon.com} \\
  \And
  Dan Goldwasser \\
  Department of Computer Science\\
  Purdue University \\
  {\tt dgoldwas@purdue.edu} \\}

\begin{document}
\maketitle
\begin{abstract}
We describe two end-to-end autoencoding models for semi-supervised graph-based projective dependency parsing. The first model is a Locally Autoencoding Parser (LAP) encoding the input using continuous latent variables in a sequential manner; The second model is a Globally Autoencoding Parser (GAP) encoding the input into dependency trees as latent variables, with exact inference. Both models consist of two parts: an encoder enhanced by deep neural networks (DNN) that can utilize the contextual information to encode the input into latent variables, and a decoder which is a generative model able to reconstruct the input. Both LAP and GAP admit a unified structure with different loss functions for labeled and unlabeled data with shared parameters. We conducted experiments on WSJ and UD dependency parsing data sets, showing that our models can exploit the unlabeled data to improve the performance given a limited amount of labeled data, and outperform a previously proposed semi-supervised model.
\end{abstract}

\section{Introduction}
\label{sec:intro}
\begin{figure}[!htp]
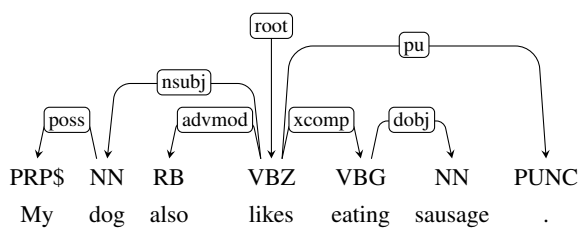

	\centering
\begin{dependency}
\begin{deptext}[column sep=.15cm, row sep=.1ex, font=\small]
PRP\$ \& NN \& RB \&[.45cm] VBZ \&[.15cm] VBG \& NN \& PUNC\\
My \& dog \& also \& likes \& eating \& sausage \& .\\
\end{deptext}
\deproot{4}{root}
\depedge{2}{1}{poss}
\depedge{4}{2}{nsubj}
\depedge{4}{3}{advmod}
\depedge{4}{5}{xcomp}
\depedge{5}{6}{dobj}
\depedge{4}{7}{pu}
\end{dependency}
	\caption{A dependency tree: directional arcs represent head-modifier relation between words.}
	\label{fig:dep-eg}
\end{figure}

Dependency parsing captures bi-lexical relationships by constructing directional arcs between words, defining a head-modifier syntactic structure for sentences, as shown in Figure \ref{fig:dep-eg}. 
Dependency trees are fundamental for many downstream tasks such as semantic parsing \citep{Reddy2016,Marcheggiani2017}, machine translation \citep{Bastings2017,Ding2007}, information extraction  \citep{Culotta2004,Liu2015} and question answering \citep{Cui2005}. As a result, efficient  parsers \citep{Kiperwasser16,Dozat2017a,Dozat2017b,Ma2018} have been developed using various neural architectures. 

Despite the success of supervised approaches, they require large amounts of labeled data, particularly when neural architectures are used. Syntactic annotation is notoriously difficult and requires specialized linguistic expertise, posing a serious challenge for low-resource languages. 
%
Semi-supervised parsing aims to alleviate this problem by combining a small amount of labeled data and a large amount of unlabeled data, to improve parsing performance over labeled data alone. Traditional semi-supervised parsers use unlabeled data to generate additional features, assisting the learning process \citep{Koo2008}, together with different variants of self-training \citep{sogaard2010simple}. However, these are usually pipe-lined and error-propagation may occur. 


In this paper, we propose two end-to-end semi-supervised parsers, illustrated in Figure \ref{fig:parsers},  namely Locally Autoencoding Parser (LAP) and Globally Autoencoding Parser (GAP). In LAP, the autoencoder model uses unlabeled examples to learn continuous latent variables of the sentence, which are then used to support tree inference by providing an enriched representation. Unlike LAP which does not perform tree inference when learning from unlabeled examples, in GAP, the dependency trees corresponding to the input sentences are treated as latent variables, and as a result the information learned from unlabeled data aligns better with the final tree prediction task.

Unfortunately, regarding trees as latent variables may cause the computation to be intractable, as the number of possible dependency trees to enumerate is exponentially large. A recent work \citep{Corro2018}, dealt with this difficulty by regenerating a particular tree of high possibility through sampling. In this study, we suggest a tractable algorithm for GAP, to directly compute all the possible dependency trees in an arc-decomposed manner, providing a tighter bound compared to  \citep{Corro2018}'s with lower time complexity. We demonstrate these advantages empirically by evaluating our model on two dependency parsing data sets.
 %
%
%
We summarize our contributions as follows:
\begin{enumerate}[noitemsep, leftmargin=*]
\item We proposed two autoencoding parsers for semi-supervised dependency parsing, with complementary strengths, trading off speed vs. accuracy; 
\item We propose a tractable inference algorithm to compute the expectation of the latent dependency tree analytically for GAP, which is naturally extendable to other tree-structured graphical models;
\item We show improved performance of both LAP and GAP with unlabeled data on WSJ and UD data sets over using labeled data alone, and over a recent semi-supervised parser \citep{Corro2018}.
\end{enumerate}



\section{Related Work}
\label{sec:work}
Most dependency parsing studies fall into two major groups: graph-based and transition-based \citep{Kubler2009}. Graph-based parsers \citep{McDonald2006} regard parsing as a structured prediction problem to find the most probable tree, while transition-based parsers \citep{Nivre2004,Nivre2008} treat parsing as a sequence of actions at different stages leading to a complete dependency tree. 

While earlier works relied on manual feature engineering, in recent years the hand-crafted features were replaced by embeddings and deep neural network architectures that were used to learn representation for scoring structural decisions, leading to improved performance in both graph-based and transition-based parsing \citep{Nivre2014,Pei2015,Chen2014,Dyer2015,Weiss2015,Andor2016,Kiperwasser16,Wiseman2016}.

The annotation difficulty for this task, has also motivated work on unsupervised (grammar induction) and semi-supervised approaches to parsing~\citep{Tu2012,Jiang2016,Koo2008,Li2014,Kiperwasser15,Cai2017,Corro2018}. It also leads to advances in using unlabeled data for constituent grammar \citep{shen2018a,shen2018b}

Similar to other structured prediction tasks, directly optimizing the objective is difficult when the underlying probabilistic model requires marginalizing over the dependency trees. Variational approaches have been widely applied to alleviating this problem, as they try to improve the lower bound of the original objective, and were employed in several recent NLP works \citep{Stratos2019,Chen2018,Kim2019a,Kim2019b}. Variational Autoencoder (VAE) \citep{Kingma2014} is particularly useful for latent representation learning, and is studied in semi-supervised context as the Conditional VAE (CVAE) \citep{Sohn2015}. Note our work differs from VAE as VAE is designed for tabular data but not for structured prediction, as in our circumstance, the input are the sentential tokens and the output is the dependency tree. 


The work mostly related to ours is \citet{Corro2018}'s as they consider the dependency tree as the latent variable, but their work takes a second approximation to the variational lower bound by an extra step to sample from the latent dependency tree, without identifying a tractable inference. We show that with the given structure, exact inference on the lower bound is achievable without approximation by sampling, which tightens the lower bound.

\section{Graph-based Dependency Parsing}
\label{sec:dparsing}

A dependency graph of a sentence can be regarded as a directed tree spanning all the words of the sentence, including a special ``word"--the ROOT--to originate out. 
Assuming a sentence of length $l$, a dependency tree can be denoted as 
$
\mathcal{T} = (<h_0, m_0>,  <h_1, m_1>,  \dots, <h_{l-1}, m_{l-1}>),
$
where $h_t$ is the index in the sequence of the head word of the dependency connecting the $t$th word $m_t$ as a modifier. 

Our graph-based parsers are constructed by following the standard structured prediction paradigm \citep{McDonald2005, Taskar2005}. In inference, based on the parameterized scoring function $\mathcal{S}_{\bm{\Lambda}}$ with parameter $\bm{\Lambda}$, the parsing problem is formulated as finding the most probable directed spanning tree for a given sentence $\bm{x}$:
\begin{equation*}
    \mathcal{T}^{*} = \arg\max_{\Tilde{\mathcal{T}} \in \mathbb{T}}\mathcal{S}_{\bm{\Lambda}}(\bm{x}, \Tilde{\mathcal{T}}),
\end{equation*}
where $\mathcal{T}^{*}$ is the highest scoring parse tree and $\mathbb{T}$ is the set of all valid trees for the sentence $\bm{x}$. 

It is common to factorize the score of the entire graph into the summation of its substructures: the individual arc scores \citep{McDonald2005}:
\begin{equation*}
    \mathcal{S}_{\bm{\Lambda}}(\bm{x}, \Tilde{\mathcal{T}}) = \sum_{(h, m)\in \Tilde{\mathcal{T}}}s_{\bm{\Lambda}}(h, m) = \sum_{t = 0}^{l-1}s_{\bm{\Lambda}}(h_t, m_t),
\end{equation*}
where $\Tilde{\mathcal{T}}$ represents the candidate parse tree, and $s_{\bm{\Lambda}}$ is a function scoring individual arcs. $s_{\bm{\Lambda}}(h, m)$ describes the likelihood of an arc from the head $h$ to its modifier $m$ in the tree. Throughout this paper, the scoring is based on individual arcs, as we focus on \textit{first order} parsing. 


\subsection{Scoring Function Using Neural Architecture}
\label{sec:neuralarc}
We used the same neural architecture as that in \citet{Kiperwasser16}'s study. We first use a bi-LSTM model to take as input $\bm{u}_{t}=[\bm{p}_{t};\bm{e}_{t}]$ at position $t$ to incorporate contextual information, by feeding the word embedding $\bm{e}_{t}$ concatenated with the POS (part of speech) tag embeddings $\bm{p}_{t}$ of each word. The bi-LSTM then projects $\bm{u}_{t}$ as $\bm{o}_{t}$.

Subsequently a nonlinear transformation is applied on these projections. Suppose the hidden states generated by the bi-LSTM are $[\bm{o}_{0}, \bm{o}_{1}, \bm{o}_{2}, \dots, \bm{o}_{t}, \dots, \bm{o}_{l-1}]$, for a sentence of length $l$, we compute the arc scores by introducing parameters $\bm{W}_{h}$, $\bm{W}_{m}$, $\bm{w}$ and $\bm{b}$, and transform them as follows:
\begin{align*}
\bm{r}_{t}^{h-arc} &= \bm{W}_{h}\bm{o}_{t};\quad \bm{r}_{t}^{m-arc} = \bm{W}_{m}\bm{o}_{t},\\
s_{\bm{\Lambda}}(h, m) &= \bm{w}^{\intercal}(\tanh(\bm{r}_{h}^{h-arc}+\bm{r}_{m}^{m-arc}+\bm{b})).
\end{align*}
In this formulation, we first use two parameters to extract two different representations that carry two different types of information: a head seeking for its modifier (h-arc); as well as a modifier seeking for its head (m-arc). Then a nonlinear function maps them to an arc score.

\begin{figure*}[htb]
\vspace{-2.4cm}
\centering
\begin{minipage}[t]{0.6\textwidth}
\centering
\begin{subfigure}{0.25\textwidth}
\centering
\includegraphics[width=1.5\textwidth]{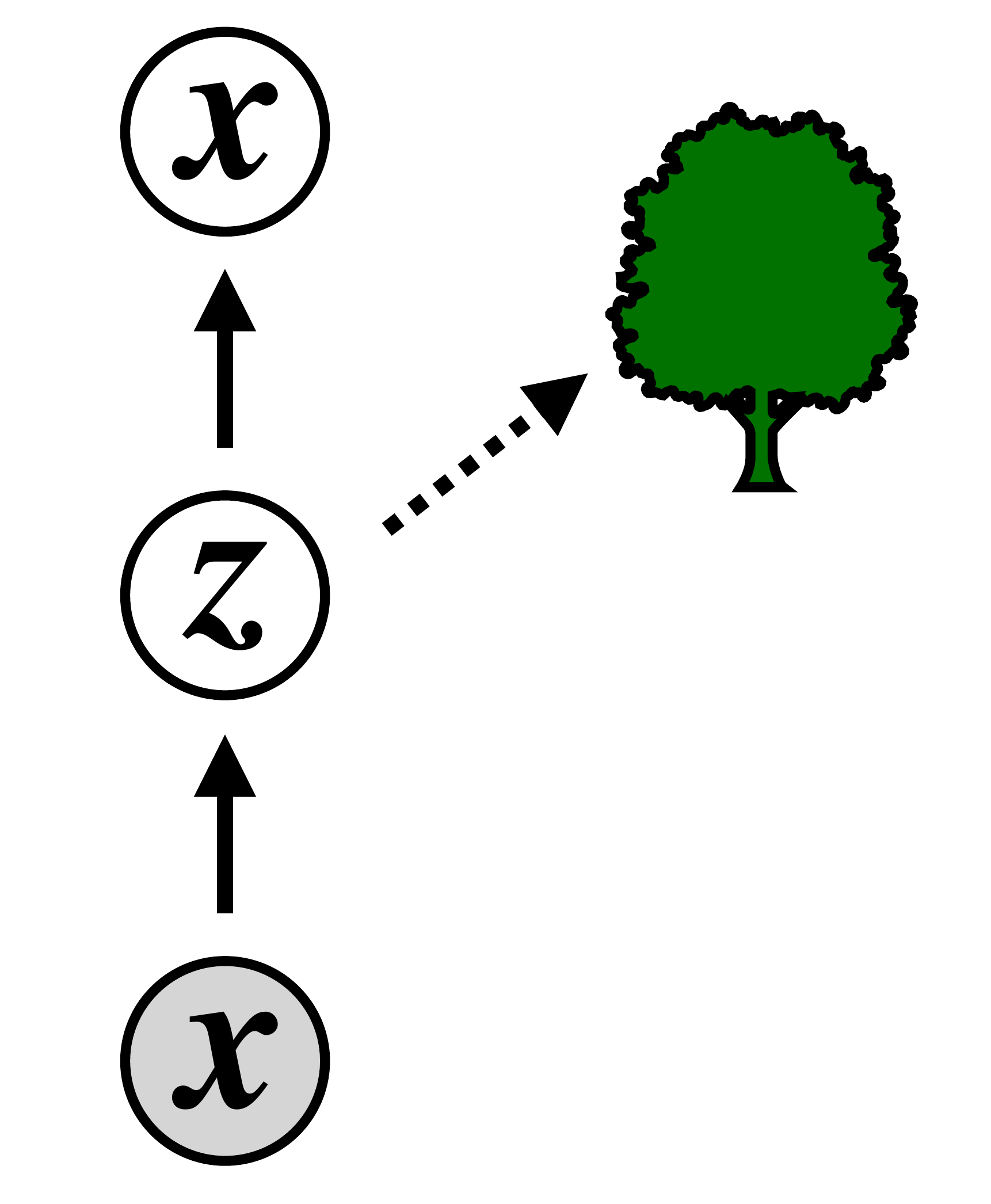}
\caption{LAP}
\label{fig:lap}
\end{subfigure}
\hspace{5em}
\begin{subfigure}{0.35\textwidth}
\centering
\includegraphics[width=1.3\textwidth]{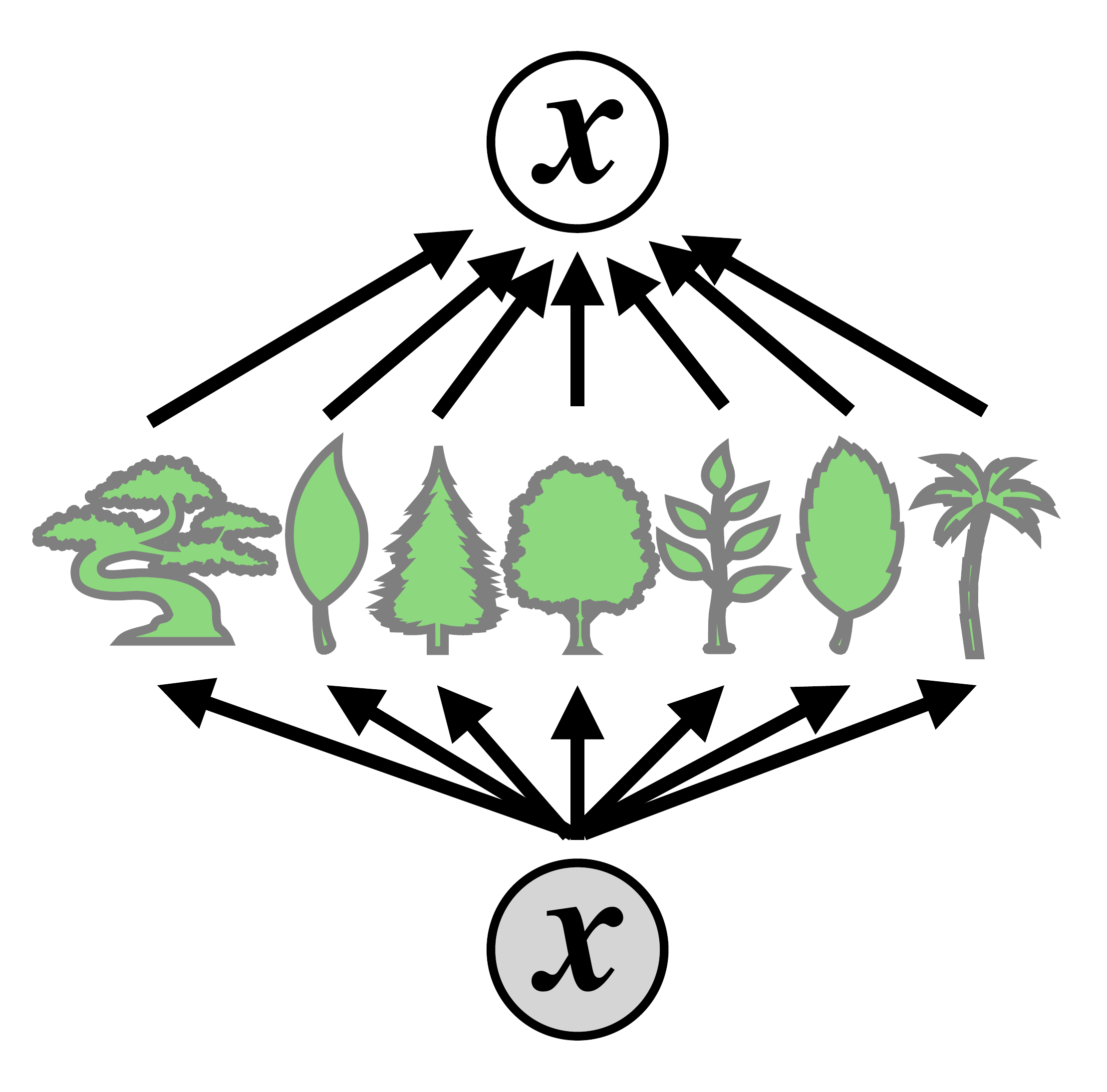}
\caption{GAP}
\label{fig:gap}
\end{subfigure}
\caption{figure}{Illustration of two different parsers. (a) LAP uses continuous latent variable to form the dependency tree (b) GAP treats the dependency tree as the latent variable.}
\label{fig:parsers}
\end{minipage}%
\hspace{0.05\textwidth}
\begin{minipage}{.3\textwidth}
\vspace{2.3cm}
\centering
\includegraphics[scale=0.23]{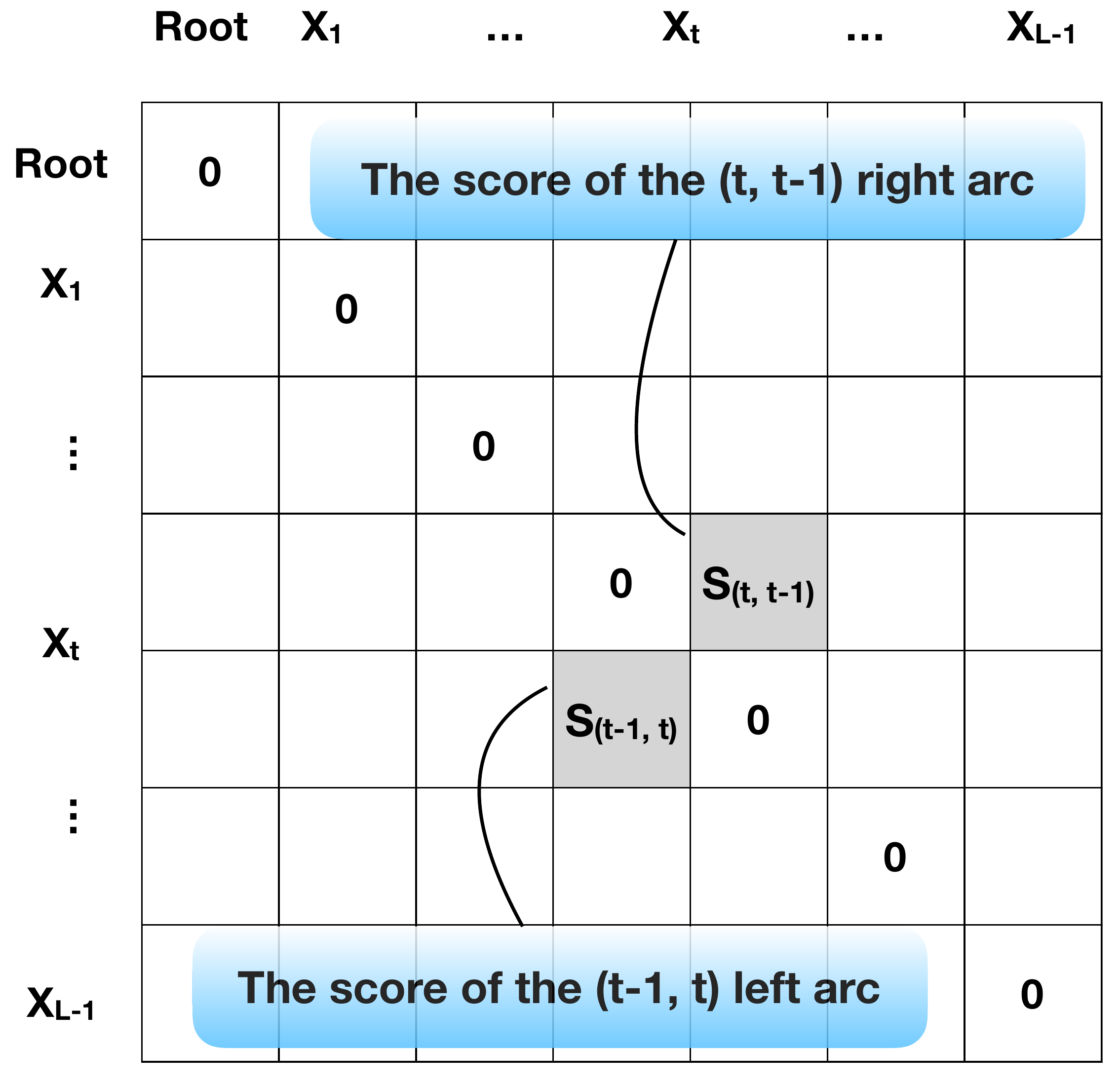}
\caption{In this illustration of the arc scoring matrix, each entry represents the $(h(head)\rightarrow m(modifier))$ score.}
\label{fig:emimat}
\end{minipage}
\end{figure*}

For a single sentence, we can form a scoring matrix as shown in Figure \ref{fig:emimat}, by filling each entry in the matrix using the score we obtained. Therefore, the scoring matrix is used to represent the head-modifier arc score for all the possible arcs connecting two tokens in a sentence \citep{Zheng2017}. 

Using the scoring arc matrix, we build graph-based parsers. Since exploring neural architectures for scoring is not our focus, we did not try other complicates, however performance shall be further improved by using advanced neural architectures~\citep{Dozat2017a, Dozat2017b}.

\section{Preliminaries}
\label{sec:pre}
To explain the LAP and GAP models, we briefly review Variational Autoencoders (VAE) and tree CRFs.
\paragraph{Variational Autoencoder (VAE)}
The typical VAE is a directed graphical model with Gaussian latent variables, denoted by $\bm{z}$. A generative process first generates latent variable $\bm{z}$ from the prior distribution $\pi(\bm{z})$ and the data $\bm{x}$ is recovered from the distribution $P_{\theta}(\bm{x}|\bm{z})$, parameterized by $\theta$. There is also an inference model $Q(\bm{z}|\bm{x})$, which is an auxiliary posterior distribution, used for inferring the most probable latent variable $\bm{z}$ given the input $\bm{x}$. In our scenario, $\bm{x}$ is an input sequence and $\bm{z}$ is a sequence of latent variables corresponding to it. 

The VAE framework seeks to maximize the complete log-likelihood $\log P(\bm{x})$ by marginalizing out the latent variable $\bm{z}$. Since direct parameter estimation of $\log P(\bm{x})$ is usually intractable, a common solution is to maximize its Evidence Lower Bound (ELBO).

\paragraph{Tree Conditional Random Field}
Linear chain CRF models an input sequence $\bm{x}=(x_{1}\dots x_{l})$ of length $l$ with labels $\bm{y}=(y_{1}\dots y_{l})$ with globally normalized probability 
\begin{align*}
    P(\bm{y}|\bm{x}) = \frac{\exp{\mathcal{S}(\bm{x}, \bm{y})}}{\sum_{\tilde{\bm{y}}\in \mathcal{Y}}\exp{\mathcal{S}(\bm{x}, \tilde{\bm{y}})}},
\end{align*}
where $\mathcal{Y}$ is the set of all the possible label sequences, and $\mathcal{S}(\bm{x}, \bm{y})$ the scoring function,  usually decomposed as emission ($\sum_{i=1}^{l}s(x_{i},y_{i})$) and transition ($\sum_{i=1}^{l}s(y_{i},y_{i+1})$) for \textit{first order} models.

Tree CRF models generalize linear chain CRF to trees. For dependency trees, if POS tags are given, the tree CRF model tries to resolve which node pairs should be connected with directed edges, such that the set of edges 
 form a tree. The potentials in the dependency tree take an exponential form, thus the conditional probability of a parse tree $\mathcal{T}$, given the sequence, can be denoted as:
\begin{equation}
  P(\mathcal{T}|\bm{x}) = \dfrac{\exp{\mathcal{S}(\bm{x}, \mathcal{T})}}{Z(\bm{x})}, \label{eq:treecrf}
\end{equation}
where $Z(\bm{x})=\sum_{\tilde{\mathcal{T}}\in\mathbb{T}(\bm{x})}\exp{\mathcal{S}(\bm{x}, \mathcal{T})}$ is the partition function that sums over all possible valid dependency trees in the set $\mathbb{T}(\bm{x})$ of the given sentence $\bm{x}$.

\section{Locally Autoencoding Parser (LAP)}
\label{sec:localmodel}

LAP is a fast, semi-supervised parser able to make use of unlabeled data in addition to labeled data. The illustration of this model is displayed in Figure \ref{fig:lap}.
LAP learns, using both labeled and unlabeled data, a continuous latent variables representation, designed to support the parsing task by creating contextualized token-representations that capture properties of the full sentence. 
Typically, each token in the sentence is represented by its latent variable $z_t$, which is a high-dimensional Gaussian variable. This configuration ensures the continuous latent variable retains the contextual information from lower-level neural models to assist finding its head or its modifier; as well as forcing the representation of similar tokens closer.
 
We adjust the original VAE setup in our semi-supervised task by considering examples with labels, similar to recent conditional variational formulations \citep{Sohn2015,Miao2016,Zhou2017}.
We propose a full probabilistic model for a given sentence $\bm{x}$, with the unified objective to maximize for both supervised and unsupervised parsing as follows:
\begin{align*}
\mathcal{J} = \log{P_{\bm{\theta}}(\bm{x})P_{\bm{\omega}}^{\epsilon}(\mathcal{T}|\bm{x})}, \quad\epsilon = \begin{cases}
    1, & \text{if}\quad \mathcal{T}\quad \text{exists},\\
    0,              & \text{otherwise}.
\end{cases}
\end{align*}

This objective can be interpreted as follows: if the training example has a golden tree $\mathcal{T}$ with it, then the objective is the log joint probability $P_{\bm{\theta}, \bm{\omega}}(\mathcal{T}, \bm{x})$; if the golden tree is missing, then the objective is the log marginal probability $P_{\bm{\theta}}(\bm{x})$. The probability of a certain tree is modeled by a tree-CRF in Eq. \ref{eq:treecrf} with parameters $\bm{\omega}$ as $P_{\bm{\omega}}(\mathcal{T}|\bm{x})$.
Given the assumed generative process $P_{\theta}(\bm{x}|\bm{z})$, directly optimizing this objective is intractable, therefore we instead optimize its Evidence Lower BOund (ELBO) :
\begin{align*}
    \mathcal{J}_{lap} = &\mathop{\mathbb{E}}\limits_{\bm{z}\sim Q_{\bm{\phi}}(\bm{z}|\bm{x})}\left[\log P_{\bm{\theta}}(\bm{x}|\bm{z})\right] -\mathbb{KL}(Q_{\bm{\phi}}(\bm{z}|\bm{x})||P_{\bm{\theta}}(\bm{z})) \\
    &+ \epsilon\mathop{\mathbb{E}}\limits_{\bm{z}\sim Q_{\bm{\phi}}(\bm{z}|\bm{x})}\left[\log P_{\bm{\omega}}(\mathcal{T}|\bm{z})\right].
\end{align*}

We show $J_{lap}$ is the ELBO of $J$ in the appendix \ref{proof:lap_elbo}. In practice, similar as VAE-style models, $\mathop{\mathbb{E}}\limits_{\bm{z}\sim Q_{\bm{\phi}}(\bm{z}|\bm{x})}\left[\log P_{\bm{\theta}}(\bm{x}|\bm{z})\right]$ is approximated by $\frac{1}{N}\sum_{j=1}^{N}\log P_{\bm{\theta}}(\bm{x}|\bm{z}_{j})$ and $\mathop{\mathbb{E}}\limits_{\bm{z}\sim Q_{\bm{\phi}}(\bm{z}|\bm{x})}\left[\log P_{\bm{\omega}}(\mathcal{T}|\bm{z})\right]$ by $\frac{1}{N}\sum_{j=1}^{N}\log P_{\bm{\omega}}(\mathcal{T}|\bm{z}_{j})$, where $\bm{z}_{j}$ is the $j$-th sample of $N$ samples sampled from $Q_{\bm{\phi}}(\bm{z}|\bm{x})$. At prediction stage, we simply use $\mu_{\bm{z}}$ rather than sampling $\bm{z}$. During the inference stage, the tree-Viterbi algorithm ensures the output is a projective dependency tree.

\section{Globally Autoencoding Parser (GAP)}
\label{sec:globalmodel}
LAP autoencodes the input locally at the sequence level, focusing on the textual representation in isolation by connecting them to the parsing algorithm. A better approach is to treat the entire dependency tree as a structured latent variable to reconstruct the input. 

We design a different model, GAP, by building a model containing both a discriminative component and a generative component to jointly learn the downstream representations and the dependency structure construction. The discriminative component builds a neural CRF model for dependency tree construction, and the generative model reconstructs the sentence from the factor graph as a Bayesian network, by assuming a generative process in which each head generates its modifier. Concretely, the latent variable in this model is the dependency tree structure.


\subsection{Discriminative Component: the Encoder}
We model the discriminative component in our model as $P_{\bm{\Phi}}(\mathcal{T}|\bm{x})$ parameterized by $\bm{\Phi}$, taking the same form as in Eq. \ref{eq:treecrf}. Typically in our model, $\bm{\Phi}$ are the parameters of the underlying neural networks, whose architecture is described in Sec. \ref{sec:neuralarc}.

\subsection{Generative Component: the Decoder}
We use a set of conditional categorical distributions to construct our Bayesian network decoder. More specifically, using the head $h$ and modifier $m$ notation, each head reconstructs its modifier with the probability $P(m_{t}|h_{t})$ for the $t$th 
word in the sentence ($0$th word is always the special ``ROOT" word), which is parameterized by the set of parameters $\bm{\Theta}$. Given $\bm{\Theta}$ as a matrix of $|\mathcal{V}|$ by $|\mathcal{V}|$, where $|\mathcal{V}|$ is the vocabulary size, $\theta_{mh}$ is the item on row $m$ column $h$ denoting the probability that the head word $h$ generates $m$. In addition, we have a simplex constraint $\sum_{m\in\mathcal{V}}\theta_{mh}=1$. The probability of reconstructing the input $\bm{x}$ as modifiers $\bm{m}$ in the 
generative process is 
\begin{equation*}
    P_{\bm{\Theta}}(\bm{m}|\mathcal{T}) = \prod_{t=0}^{l-1}P(m_{t}|h_{t}) = \prod_{t=0}^{l-1}\theta_{m_{t}h_{t}},
\end{equation*}
where $l$ is the sentence length and $P(m_{t}|h_{t})$ represents the probability a head generating its modifier. 

\subsection{A Unified Supervised and Unsupervised Learning Framework }
With the design of the discriminative component and the generative component of the proposed model, we have a unified learning framework for sentences with or without golden parse tree.

The complete data likelihood of a given sentence, if the golden tree is given, is 
\begin{align*}
        P_{\bm{\Theta}, \bm{\Phi}}(\bm{m}, \mathcal{T}|\bm{x}) = & P_{\bm{\Theta}}(\bm{m}|\mathcal{T})P_{\bm{\Phi}}(\mathcal{T}|\bm{x})\\
        = & \left[\prod_{t=1}^{l}P(m_{t}|h_{t})\right] \dfrac{\exp{\mathcal{S}_{\bm{\Phi}}(\bm{x}, \mathcal{T})}}{Z(\bm{x})}\\
        = & \dfrac{\exp{\sum\limits_{(h, m)\in\mathcal{T}}s^{'}_{\bm{\Phi},\bm{\Theta}}(h, m)}}{Z(\bm{x})},
\end{align*}
where $s^{'}_{\bm{\Phi},\bm{\Theta}}(h, m) = s_{\bm{\Phi}}(h, m) + \log\theta_{mh}$, with $\bm{m}, \bm{x}$ and $\mathcal{T}$ all observable.

For a unlabeled sentence, the complete data likelihood can be obtained by marginalizing over all the possible parse trees in the set $\mathbb{T}(\bm{x})$:
\begin{align*}
    P_{\bm{\Theta}, \bm{\Phi}}(\bm{m}|\bm{x}) = &\sum_{\mathcal{T} \in \mathbb{T}(\bm{x})} P_{\bm{\Theta}, \bm{\Phi}}(\bm{m}, \mathcal{T}|\bm{x})
    \\
    =&\dfrac{U(\bm{x})}{Z(\bm{x})},
\end{align*}
where $U(\bm{x}) = \sum_{\mathcal{T} \in \mathbb{T}(\bm{x})}\exp{\sum\limits_{(h, m)\in\mathcal{T}}s^{'}_{\bm{\Phi},\bm{\Theta}}(h, m)} $.


We adapted a variant of \citet{Eisner1996}'s algorithm  to marginalize over all possible trees to compute both $Z$ and $U$, as $U$ has the same structure as $Z$, assuming a projective tree. This algorithm also enforces the output is a projective dependency tree.

We use log-likelihood as our objective function. The objective for a sentence with golden tree is:
\begin{align*}
    \mathcal{J}_{l} =& \log P_{\bm{\Theta}, \bm{\Phi}}(\bm{m}, \mathcal{T}|\bm{x})\\
    =& \sum\limits_{(h, m)\in\mathcal{T}}s^{'}_{\bm{\Phi},\bm{\Theta}}(h, m) - \log Z(\bm{x})
\end{align*}
If the input sentence does not have an annotated golden tree, then the objective is:
\begin{align}
    \mathcal{J}_{u} =& \log P_{\bm{\Theta}, \bm{\Phi}}(\bm{m}|\bm{x})\notag
    \\
    =& \log U(\bm{x}) - \log Z(\bm{x}).\label{lossU}
\end{align}
Thus, during training, the objective function with shared parameters is chosen based on whether the sentence in the corpus has golden parse tree or not. 

\subsection{Learning}
Directly optimizing the loss in Eq.\ref{lossU} is difficult when using the unlabeled data, and may lead to undesirable shallow local optima. Instead, we derive the evidence lower bound (ELBO) of $\log P_{\bm{\Theta}, \bm{\Phi}}(\bm{m}|\bm{x})$ as follows, by denoting $Q(\mathcal{T}) = P_{\bm{\Theta}, \bm{\Phi}}(\mathcal{T} |\bm{m}, \bm{x})$ as the posterior:
\begin{align*}
    \log P_{\bm{\Theta}, \bm{\Phi}}(\bm{m}|\bm{x}) &= \log \sum_{\mathcal{T}}Q(\mathcal{T})\dfrac{P_{\bm{\Theta}, \bm{\Phi}}(\bm{m}, \mathcal{T}|\bm{x})}{Q(\mathcal{T})}\\
    &=\log \mathbb{E}_{\mathcal{T}\sim Q(\mathcal{T})}\dfrac{P_{\bm{\Theta}, \bm{\Phi}}(\bm{m}, \mathcal{T}|\bm{x})}{Q(\mathcal{T})}\\
    &\geq \mathbb{E}_{\mathcal{T}\sim Q(\mathcal{T})}\log\dfrac{P_{\bm{\Theta}, \bm{\Phi}}(\bm{m}, \mathcal{T}|\bm{x})}{Q(\mathcal{T})}\\
    &= \mathbb{E}_{\mathcal{T}\sim Q(\mathcal{T})}\left[\log P_{\bm{\Theta}}(\bm{m}|\mathcal{T})\right] \\
    &- \mathbb{KL}\left[Q(\mathcal{T})||P_{\bm{\Phi}}(\mathcal{T}|\bm{x})\right].
\end{align*}

Instead of maximizing the log-likelihood directly, we alternatively maximize the ELBO, so our new objective function for unlabeled data becomes
\begin{align*}
    \max_{\bm{\Theta}, \bm{\Phi}}&\mathbb{E}_{\mathcal{T}\sim Q(\mathcal{T})}\left[\log P_{\bm{\Theta}}(\bm{m}|\mathcal{T})\right] - \mathbb{KL}\left[Q(\mathcal{T})||P_{\bm{\Phi}}(\mathcal{T}|\bm{x})\right].\label{eq:gapelbo}
\end{align*}
Note instead of using sampling to approximate the ELBO above as in \cite{Corro2018}'s work: $\mathbb{E}_{\mathcal{T}\sim Q(\mathcal{T})}\left[\log P_{\bm{\Theta}}(\bm{m}|\mathcal{T})\right] - \mathbb{KL}\left[Q(\mathcal{T})||P_{\bm{\Phi}}(\mathcal{T}|\bm{x})\right] \approx \frac{1}{N}\sum_{j=1}^{N}\left[\log P_{\bm{\Theta}}(\bm{m}|\mathcal{T}_{j})\right] - \mathbb{KL}\left[Q(\mathcal{T}_{j})||P_{\bm{\Phi}}(\mathcal{T}_{j}|\bm{x})\right]$, we identify a tractable algorithm to calculate it directly, tightening the bound.

In addition, to account for the unambiguity in the posterior, we incorporate entropy regularization \citep{Tu2012} when applying our algorithm, by adding an entropy term $-\sum_{\mathcal{T}}Q(\mathcal{T})\log Q(\mathcal{T})$ with a non-negative factor $\sigma$ when the input sentence does not have a golden tree. Adding this regularization term is equivalent as raising the expectation of $Q(\mathcal{T})$ to the power of $\frac{1}{1-\sigma}$. We annealed $\sigma$ from the beginning of training to the end, as in the beginning, the generative model is well initialized by sentences with golden trees that resolve disambiguity. The details of training are shown in Alg. \ref{algo:vl}.

\begin{figure}[tb]
  \centering
  	\includegraphics[scale=0.25]{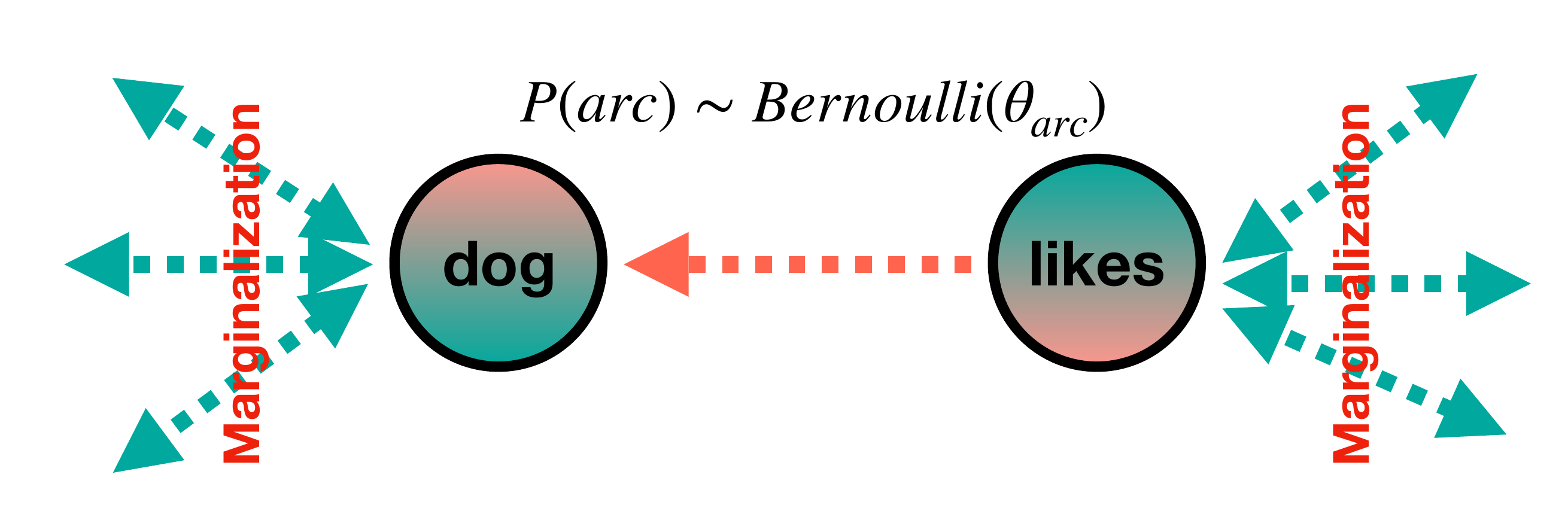}
	\caption{We illustrate the tractable inference procedure by marginalization in a arc-decomposed manner.}
	\label{fig:arc}
\end{figure}

\begin{algorithm}[!t]
\caption{Learning Algorithm for GAP}
\label{algo:vl}
\begin{algorithmic}[1]
\State{Initialize the parameter $\bm{\Theta}$ in the decoder with the labeled data set $ \{\bm{x}, \mathcal{T}\}^{l} $.}
\State{Initialize $\bm{\Lambda}$ in the encoder randomly.}
\For{$t$ in $epochs$}
\For{sentence $\bm{x}_{i}^{l}$ with golden parse tree $\mathcal{T}_{i}^{l}$ in the labeled data set $ \{\bm{x}, \mathcal{T}\}^{l} $}
\State{Stochastically update the parameter $\bm{\Lambda}$ in the encoder using Adam while fixing the decoder.}
\EndFor
\State{Initialize a Counting Buffer $\mathcal{B}$}
\For{unlabeled sentence $\bm{x}_{i}^{u}$ in the unlabeled data set $ \{\bm{x}\}^{u} $}
\State{Compute the posterior $Q(\mathcal{T})$ in an arc factored manner for $\bm{x}_{i}^{u}$ tractably.}
\State{Compute the expectation of all possible $(h(head)\rightarrow m(modifier))$ occurrence in the sentence $\bm{x}$ based on $Q(\mathcal{T})$.}
\State{Update buffer $\mathcal{B}$ using the expectation to the power for $\frac{1}{1-\sigma}$ of all possible $(h\rightarrow m)$.}
\EndFor
\State{Obtain $\bm{\Theta}$ globally and analytically based on the buffer $\mathcal{B}$ and renew the decoder.}
\EndFor
\end{algorithmic}
\end{algorithm}
\vspace{-0.5em}
\subsection{Convexity of ELBO w.r.t. $\bm{\Theta}$}
In practice, we found the model benefits more by fixing the parameter $\bm{\Phi}$ when the data is unlabeled and optimizing the ELBO w.r.t. the parameter $\bm{\Theta}$. We attribute this to the strict convexity of the ELBO w.r.t. $\bm{\Theta}$, by sketching the proof in Appendix \ref{proof:convexity_gap}.


\subsection{Tractable Inference}
The common approach to approximate the expectation of the latent variables from the posterior distribution $Q(\mathcal{T})$ is via sampling in VAE-type models \citep{Kingma2014}. In a significant contrast to that, we argue in this GAP model the expectation of the latent variable (which is the dependency tree structure) is analytically tractable by designing a variant of the inside-outside algorithm \citep{Eisner1996, Paskin2001} in an arc decomposed manner. This can be done by regarding each directed arc as an indicator variable from a Bernoulli distribution, implying whether the arc exists or not. 
We argue that assuming the dependency tree is \textit{projective}, specialized \textit{belief propagation} algorithm exists to compute not only the \textit{marginalization} over all other arcs but also the target arc. This algorithm also computes the \textit{expectation} of each arc analytically, making inference tractable. This idea is illustrated in Figure \ref{fig:arc}. We show the details of this algorithm in the appendix \ref{full algo}. We also show how to calculate the expectation w.r.t the posterior distribution $Q(\mathcal{T})$ in appendix \ref{sec:expectation}.




\section{Experiments}
\label{sec:exp}

\subsection{Experimental Settings}

\paragraph{Data sets}
First we compared our models' performance with strong baselines on the WSJ data set, which is the Stanford Dependency conversion \citep{DeMarneffe2008} of the Penn Treebank \citep{Marcus1993} using the standard section split: 2-21 for training, 22 for development and 23 for testing. Second we evaluated our models on multiple languages, using data sets from UD (Universal Dependency) 2.3 \citep{UD}. 
Since semi-supervised learning is particularly useful for low-resource languages, we believe those languages in UD can benefit from our approach. The statistics of the data used in our experiments are described in Table \ref{tbl:ud_stat}. 

To simulate the low-resource language environment, we used $10\%$ of the whole training set as the annotated, and the rest $90\%$ as the unlabeled.


\begin{table*}[!htb]
\centering
\small
\begin{tabular}{|l|c|c|c|c|c|c|c|c|c|c|c|c|}
\hline 
Language & WSJ & Dutch & Spanish & English & French & Croatian & German & Italian & Russian & Japanese\\
\hline
Training & 39832 & 12269 & 14187 & 2914 & 14450 & 6983 & 13814 & 13121 & 3850 & 7133 \\
\hline
Development & 1700 & 718 & 1400 & 707 & 1476 & 849 & 799 & 564 & 579 & 511 \\
\hline
Testing & 2416 & 596 & 426 & 769 & 416 & 1057 & 977 & 482 & 601 & 551 \\
\hline
\end{tabular}
\caption{Statistics of multiple languages we used in our experiments are shown here. The table shows number of sentences in the training, development and test data divisions.}
\label{tbl:ud_stat}
\end{table*}

\begin{table*}[!htb]
\centering
\small
\begin{tabular}{|l|c|c|c|c|c|c|c|c|c|c|c|}
\hline 
Model & Dutch & Spanish & English & French & Croation & German & Italian & Russian & Japanese\\
\hline
NMP (L) & 76.11 & 82.00 & 75.51 & 83.07 & 77.44 & 74.07 & 82.85 & 75.18 & 93.46 \\
\hline
NTP (L) & 76.20 & 82.09 & 75.57 & 83.12 & 77.51 & 74.13 & 82.99 & 75.23 & 93.54 \\
\hline
LAP (L) & 76.15 & 81.93 & 75.36 & 83.09 & 77.45 & 74.14 & 83.07 & 74.84 & 93.38 \\
\hline
GAP (L) & 76.23 & 81.97 & 75.75 & 83.11 & 77.49 & 74.16 & 83.14 & 75.17 & 93.52 \\
\hline
\hline
CRFAE (L+U) & 71.32 & 74.67 & 68.52 & 77.35 & 69.89 & 68.44 & 76.37 & 68.64  & 87.26\\
\hline
ST (L+U) & 75.37 & 80.86 & 72.76 & 81.38 & 76.10 & 73.45 & 82.74 & 72.57 & 91.43 \\
\hline
LAP (L+U) & 76.29 & 82.48 & 75.48 & 83.23 & 77.78 & 74.48 & 83.34 & 75.22 & 93.65 \\
\hline
GAP (L+U) & 76.54 & 82.56 & 76.21 & 83.26 & 77.83 & 74.63 & 83.54 & 75.69 & 93.92\\
\hline
\end{tabular}
\caption{In this table we compare different models on multiple languages from UD. Models were trained in a fully supervised fashion with labeled data only (noted as ``L") or semi-supervised (notes as ``L+U"). ``ST" stands for self-training.}
\label{tbl:udresults}
\vspace{-1em}
\end{table*}

\paragraph{Training}
In the training phase, we use Adam \citep{Kingma2014b} to update all the parameters in both LAP and GAP, except the parameters in the decoder in GAP, which are updated by using their global optima in each epoch. In GAP, We annealed $\sigma$ from $1$ to $0.3$.
We did not take efforts to tune models' hyper-parameters and they remained the same across all the experiments.
To preventing over-fitting, we applied the ``early stop" strategy by using the development set.

\subsection{Semi-Supervised Dependency Parsing on WSJ Data Set}
We first evaluate our models on the WSJ data set and compared the model performance with other semi-supervised parsing models, including CRFAE \citep{Cai2017}, which is originally designed for dependency grammar induction but can be modified for semi-supervised parsing, and ``differentiable Perturb-and-Parse" parser (DPPP) \citep{Corro2018}. To contextualize the results, we also experiment with the supervised neural margin-based parser (NMP) \citep{Kiperwasser16}, neural tree-CRF parser (NTP) and the supervised version of LAP and GAP, with only the labeled data. 
To ensure a fair comparison, our experimental set up on the WSJ is identical as that in DPPP, using the same 100 dimension skip-gram word embeddings employed in an earlier transition-based system \citep{Dyer2015}. We show our experimental results in Table~\ref{tbl:wsj_res}.

\begin{table}[!ht]
\centering
\small
\begin{tabular}{|l|c|}
\hline 
Model & UAS\\
\hline
 DPPP\citep{Corro2018}(L) & 88.79 \\
 \hline
 DPPP\citep{Corro2018}(L+U) & 89.50 \\
\hline
 CRFAE\citep{Cai2017}(L+U) & 82.34 \\
\hline
 NMP\citep{Kiperwasser16}(L) & 89.64 \\
 \hline
 NTP (L) & 89.63 \\
 \hline
 Self-training (L+U) & 87.81\\
 \hline
 LAP (L) & 89.37\\
 \hline
 LAP (L+U) & \textbf{89.49}\\
 \hline
 GAP (L) & 89.65\\
 \hline
 GAP (L+U) & \textbf{89.96}\\
\hline
\end{tabular}
\caption{Comparing model performance on WSJ data set with $10\%$ labeled data. ``L" means only $10\%$ labeled data is used, while ``L+U" means both $10\%$ labeled and $90\%$ unlabeled data are used. 
}
\label{tbl:wsj_res}
\vspace{-2em}
\end{table}

As shown in this table, both of our LAP and GAP model are able to utilize the unlabeled data to increase the overall performance comparing with only using labeled data. Our LAP model performs slightly worse than the NMP model, which we attribute to the increased model complexity by incorporating extra encoder and decoders to deal with the latent variable. However, our LAP model achieved comparable results on semi-supervised parsing as the DPPP model, while our LAP model is simple and straightforward without additional inference procedure. Instead, the DPPP model has to sample from the posterior of the structure by using a ``GUMBEL-MAX trick" to approximate the categorical distribution at each step, which is intensively computationally expensive. Further, our GAP model achieved the best results among all these methods, by successfully leveraging the the unlabeled data in an appropriate manner. We owe this success to such a fact: GAP is able to calculate the exact expectation of the arc-decomposed latent variable, \textit{the dependency tree structure}, in the ELBO for the complete data likelihood when the data is unlabeled, rather than using sampling to approximate the true expectation. Self-training using NMP with both labeled and unlabeled data is also included as a base-line, where the performance is deteriorated without appropriately using the unlabeled data.


\subsection{Semi-supervised Dependency Parsing on the UD Data Set}

We  also  evaluated  our  models  on multiple languages from the UD data and compared the performance with the semi-supervised version of CRFAE and the fully supervised NMP and NTP. To fully simulate the low-resource scenario, we did not use any external word embeddings but initializing them randomly.

We summarize the results in Table \ref{tbl:udresults}. First, when using labeled data only, LAP and GAP have similar performance as NMP and NTP. Second, we note that our LAP and GAP models do benefit from the unlabeled data, compared to using labeled data only. Both our LAP and GAP model are able to exploit the hidden information in the unlabeled data to improve the performance. Comparing between LAP and GAP, we notice GAP in general has better performance than LAP, and can better leverage the information in the unlabeled data to boost the performance. These results validate that GAP is especially useful for low-resource languages with few annotations. We also experimented using self-training on the labeled and unlabeled data with the NMP model. As results show, self-training deteriorate the performance especially when the size of the training data is small.


\section{Comparison of Complexity}
\label{sec:complexity}
We briefly compare the complexity of LAP, GAP and their competitors in
Table \ref{tbl:complexity}. As can be seen, though all the algorithms are of $O(l^3)$ complexity, the constant differs significantly. Our LAP model is 6 times faster than the DPPP algorithm while our GAP model is 2 times faster. Considering the model performance, our models are favored.

\begin{table}[!ht]
\centering
\small
\begin{tabular}{|l|c|c|}
\hline 
Model & Complexity (Train) & Complexity (Eval)\\
\hline
 DPPP & 24$l^3$ & 4$l^3$ \\
 \hline
 CRFAE & 12$l^3$ & 4$l^3$ \\
\hline
 NMP\footnotemark & 4$l^3$ & 4$l^3$ \\
 \hline
 NTP & 4$l^3$ & 4$l^3$ \\
 \hline
 LAP & 4$l^3$ & 4$l^3$ \\
 \hline
 GAP & 12$l^3$ & 4$l^3$ \\
 \hline
\end{tabular}
\caption{We compare the complexity of different models, with $l$ indicating the sentence length.
}
\label{tbl:complexity}
\vspace{-1.5em}
\end{table}
\footnotetext{Although it is the same as NTP and LAP, in fact the \textit{max} operation is faster than the \textit{sum} operation.}

\section{Conclusion}
\label{sec:concl}
In this paper, we present two semi-supervised parsers, namely locally autoencoding parser (LAP) and globally autoencoding parser (GAP). Both are end-to-end learning systems enhanced with neural architectures, capable of utilizing the latent information within the unlabeled data together with labeled data to improve the parsing performance, without using external resources. More importantly, GAP outperforms the previous published \citep{Corro2018} semi-supervised parsing system on the WSJ data set. We attribute this success to two reasons: First, GAP consists both a discriminative component and a generative component, which are constraining and supplementing each other such that final parsing choices are made in a checked-and-balanced manner to avoid over-fitting. Second, instead of sampling from posterior of the latent variable (the dependency tree), GAP analytically computes the expectation and marginalization, hence the decoder's global optima can be found, leading to improved performance.

\section{Acknowledgements}
We sincerely appreciate Prof. Kewei Tu for his valuable inspiration, comments, discussion and suggestions. We also thank Ge Wang, Dingquan Wang, Jiong Cai for their constructive discussion. We extend our thanks to all the anonymous reviewers for their insightful feedback. This work was partially funded by an Nvidia GPU grant. 

\bibliography{mybib}
\bibliographystyle{named}

\clearpage
\appendix
\section{Appendix}
\label{sec:appendix}

\subsection{ELBO of LAP's Original Ojective}
\label{proof:lap_elbo}
Given an input sequence $\bm{x}$, we prove that $\mathcal{J}_{lap}$ is the ELBO of $\mathcal{J}$:
\begin{lemma} $\mathcal{J}_{lap}$ is the ELBO (evidence lower bound) of the original objective $\mathcal{J}$, with an input sequence $\bm{x}$.
\label{lemma:lapelbo}
\end{lemma}
Denote the encoder $Q$ is a distribution used to approximate the true posterior distribution $P_{\bm{\phi}}(\bm{z}|\bm{x})$, parameterized by $\bm{\phi}$ such that $Q$ encoding the input into the latent space $\bm{z}$. 
\begin{proof}
\begin{align*}
\log{P_{\bm{\theta}}(\bm{x})P_{\bm{\omega}}^{\epsilon}(\mathcal{T}|\bm{x})} =& \underbrace{\log{P_{\bm{\theta}}(\bm{x})}}_{\mathcal{U}} + \underbrace{\epsilon\log{P_{\bm{\omega}}(\mathcal{T}|\bm{x})}}_{\mathcal{L}}\\
\mathcal{U} =& \log \int_{z}Q_{\bm{\phi}}(\bm{z}|\bm{x})\dfrac{P_{\bm{\theta}}(\bm{x})}{Q_{\bm{\phi}}(\bm{z}|\bm{x})}d_{z}\\
\geq& \mathop{\mathbb{E}}\limits_{\bm{z}\sim Q_{\bm{\phi}}(\bm{z}|\bm{x})}\left[\log P_{\bm{\theta}}(\bm{x}|\bm{z})\right]\\
-&\mathop{\mathbb{E}}\limits_{\bm{z}\sim Q_{\bm{\phi}}(\bm{z}|\bm{x})}\left[\log \dfrac{Q_{\bm{\phi}}(\bm{z}|\bm{x})}{P_{\bm{\theta}}(\bm{x})}\right]\\
=&\mathop{\mathbb{E}}\limits_{\bm{z}\sim Q_{\bm{\phi}}(\bm{z}|\bm{x})}\left[\log P_{\bm{\theta}}(\bm{x}|\bm{z})\right]\\
-&\mathbb{KL}\left(Q_{\bm{\phi}}(\bm{z}|\bm{x})||P_{\bm{\theta}}(\bm{z})\right), \\
&\text{(ELBO of traditional VAE)}\\
\mathcal{L} =& \epsilon\log{P_{\bm{\omega}}(\mathcal{T}|\bm{x})} \\
=& \epsilon\log{ \int_{z}P_{\bm{\omega}}(\mathcal{T}|\bm{z})Q_{\bm{\phi}}(\bm{z}|\bm{x})d_{z} }\\
=& \epsilon\log{\mathop{\mathbb{E}}\limits_{\bm{z}\sim Q_{\bm{\phi}}(\bm{z}|\bm{x})}\left[ P_{\bm{\omega}}(\mathcal{T}|\bm{z})\right]}\\
\geq& \epsilon\mathop{\mathbb{E}}\limits_{\bm{z}\sim Q_{\bm{\phi}}(\bm{z}|\bm{x})}\left[\log P_{\bm{\omega}}(\mathcal{T}|\bm{z})\right].
\end{align*}
Combining $\mathcal{U}$ and $\mathcal{L}$ leads to the fact:
\begin{align*}
\mathcal{U} + \mathcal{L} \geq& \mathop{\mathbb{E}}\limits_{\bm{z}\sim Q_{\bm{\phi}}(\bm{z}|\bm{x})}\left[\log P_{\bm{\theta}}(\bm{x}|\bm{z})\right] -\mathbb{KL}(Q_{\bm{\phi}}(\bm{z}|\bm{x})||P_{\bm{\theta}}(\bm{z})) \\
+& \epsilon\mathop{\mathbb{E}}\limits_{\bm{z}\sim Q_{\bm{\phi}}(\bm{z}|\bm{x})}\left[\log P_{\bm{\omega}}(\mathcal{T}|\bm{z})\right] = \mathcal{J}_{lap}
\end{align*}
\end{proof}

\subsection{Convexity of ELBO w.r.t $\bm{\Theta}$ in GAP}
\label{proof:convexity_gap}
Since we only care about the term containing $\bm{\Theta}$, the KL divergence term degenerates to a constant. For sentence $i$, $Q(\mathcal{T}_{i})$ has been derived in the previous section as matrix $\bm{P}$ and $\mathbbm{1}$ is the indication function.

\begin{align}
\max_{\bm{\Theta}}&\sum\limits_{i}\mathbb{E}_{\mathcal{T}_{i}\sim Q(\mathcal{T}_{i})}\left[\log P_{\bm{\Theta}}(\bm{m}_{i}|\mathcal{T}_{i})\right]\notag\\ 
&- \mathbb{KL}\left[Q(\mathcal{T}_{i})||P_{\bm{\Phi}}(\mathcal{T}_{i}|\bm{x}_{i})\right] \notag\\
\max_{\bm{\Theta}}& \sum\limits_{i}\sum\limits_{\mathcal{T}_{i}\in \mathbb{T}(\bm{x_{i}})}Q(\mathcal{T}_{i})\log P_{\bm{\Theta}}(\bm{m}_{i}|\mathcal{T}_{i}) + Const \notag \\
\max_{\bm{\Theta}}& \sum\limits_{(h\rightarrow m)}\log \theta_{mh}\mathrm{E}_{(h\rightarrow m) \sim Q}\mathbbm{1}(h\rightarrow m)\notag \\
\max_{\bm{\Theta}}& \sum\limits_{(h\rightarrow m)}Q(\mathbbm{1}(h\rightarrow m))\log \theta_{mh},\\
&\text{$Q(\mathbbm{1}(h\rightarrow m))$ is a Bernoulli distribution,}\notag\\ 
&\text{indicating whether the arc $(h\rightarrow m)$ exists.}\notag\\
s.t. &\sum\limits_{m}\theta_{mh} = 1\quad \forall h. 
\end{align}

\subsection{Marginalization and Expectation of Latent Parse Trees}
\label{sec:expectation}
Light modification is needed in our study to calculate the expectation w.r.t. the posterior distribution $Q(\mathcal{T}) = P_{\bm{\Theta}, \bm{\Phi}}(\mathcal{T} |\bm{m}, \bm{x})$, as we have 

\begin{align*}
    &P_{\bm{\Theta}, \bm{\Phi}}(\mathcal{T} |\bm{m}, \bm{x}) = \dfrac{P_{\bm{\Theta}, \bm{\Phi}}(\mathcal{T}, \bm{m} |\bm{x})}{P_{\bm{\Theta}, \bm{\Phi}}(\bm{m} |\bm{x})}&\\
    &= \dfrac{\exp^{\sum\limits_{(h, m)\in\mathcal{T}}s^{'}_{\bm{\Phi},\bm{\Theta}}(h, m)}}{Z(\bm{x})}/\sum_{\mathcal{T} \in \mathbb{T}}\left[\dfrac{\exp^{\sum\limits_{(h, m)\in\mathcal{T}}s^{'}_{\bm{\Phi},\bm{\Theta}}(h, m)}}{Z(\bm{x})}\right]&\\
     &= \dfrac{\exp{\sum\limits_{(h, m)\in\mathcal{T}}s^{'}_{\bm{\Phi},\bm{\Theta}}(h, m)}}{Z'(\bm{x})}&,
\end{align*}
where $Z'(\bm{x}) = \sum_{\mathcal{T} \in \mathbb{T}}\exp{\sum\limits_{(h, m)\in\mathcal{T}}s^{'}_{\bm{\Phi},\bm{\Theta}}(h, m)}$ is the real marginal we need to calculate using the transformed scoring matrix $\bm{S}'$ as input in the inside algorithm. Each entry in this transformed scoring matrix is defined in the text as $s^{'}_{\bm{\Phi},\bm{\Theta}}(h, m)$. 

\subsection{Algorithm Details for GAP}
\label{full algo}
Assuming the sentence is of length $l$, we could obtain an arc decomposed scoring matrix $\bm{S}$ of size $l\times l$, with the entry $\bm{S}[i,j]_{i\neq j,j\neq 0}$ stands for the arc score where $i$th word is the head and $j$th word the modifier.

We first describe the \textbf{inside} algorithm to compute the marginalization of all possible projective trees in Algo.\ref{algo:inside}.

\begin{algorithm*}[!htp]
\caption{Inside Algorithm}
\label{algo:inside}
\hspace*{\algorithmicindent} \textbf{Input: $\bm{S}$} \\
\hspace*{\algorithmicindent} \textbf{Output: $\bm{\alpha},Z$}
\begin{algorithmic}[1]
\State{$\bm{\alpha}\leftarrow -\infty$}
\For{$s \in 0\dots l-1$}
\If{$s>0$}
\State{$\bm{\alpha}[s,s,L,C] \leftarrow 0$} 
\EndIf
\State{$\bm{\alpha}[s,s,R,C] \leftarrow 0$} 
\EndFor
\For{$k \in 1\dots l-1$}
\For{$s \in 0\dots l-k$}
\State{$t=s+k$}
\If{$s>0$}
\State{$\bm{\alpha}[s,t,L,I] \leftarrow \log\sum\limits_{u\in[s,t-1]}\exp\left(\bm{\alpha}[s,u,R,C]+\bm{\alpha}[u+1,t,L,C]\right)+\bm{S}[t,s]$}
\State{\begin{tikzpicture}
\draw [thick] (-1,0) to [round left paren] (-1,1);
\draw[fill=gray!30] (1,0) -- (0,0) -- (0,0.5) -- (1,1) -- cycle;
\node at (2,0.5) {$=$};
\draw (4,0) -- (3,0) -- (3,1) -- cycle;
\draw [<-, thick] (3.25,1) to [out=30,in=150] (5.75,1); 
\draw (5,0) -- (6,0) -- (6,1) -- cycle;
\draw [thick] (7,0) to [round right paren] (7,1);
\end{tikzpicture}}
\EndIf
\State{$\bm{\alpha}[s,t,R,I] \leftarrow \log\sum\limits_{u\in[s,t-1]}\exp\left(\bm{\alpha}[s,u,R,C]+\bm{\alpha}[u+1,t,L,C]\right)+\bm{S}[s,t]$}
\State{\begin{tikzpicture}
\draw [thick] (-1,0) to [round left paren] (-1,1);
\draw[fill=gray!30] (1,0) -- (0,0) -- (0,1) -- (1,0.5) -- cycle;
\node at (2,0.5) {$=$};
\draw (4,0) -- (3,0) -- (3,1) -- cycle;
\draw [->, thick] (3.25,1) to [out=30,in=150] (5.75,1); 
\draw (5,0) -- (6,0) -- (6,1) -- cycle;
\draw [thick] (7,0) to [round right paren] (7,1);
\end{tikzpicture}}
\If{$s>0$}
\State{$\bm{\alpha}[s,t,L,C] \leftarrow \log\sum\limits_{u\in[s,t-1]}\exp\left(\bm{\alpha}[s,u,L,C]+\bm{\alpha}[u,t,L,I]\right)$}
\State{\begin{tikzpicture}
\draw [thick] (-1,0) to [round left paren] (-1,1);
\draw[fill=gray!30] (1,0) -- (0,0) -- (1,1) -- cycle;
\node at (2,0.5) {$=$};
\draw (4,0) -- (3,0) -- (4,0.5) -- cycle;
\draw [-, thick] (4,0.5) -- (5,0.5); 
\draw (5,0) -- (6,0) -- (6,1) -- (5,0.5) -- cycle;
\draw [thick] (7,0) to [round right paren] (7,1);
\end{tikzpicture}}
\EndIf
\State{$\bm{\alpha}[s,t,R,C] \leftarrow \log\sum\limits_{u\in[s,t-1]}\exp\left(\bm{\alpha}[s,u+1,R,I]+\bm{\alpha}[u+1,t,R,C]\right)$}
\State{\begin{tikzpicture}
\draw [thick] (-1,0) to [round left paren] (-1,1);
\draw[fill=gray!30] (1,0) -- (0,0) -- (0,1) -- cycle;
\node at (2,0.5) {$=$};
\draw (4,0) -- (3,0) -- (3,1) -- (4,0.5) -- cycle;
\draw [-, thick] (4,0.5) -- (5,0.5); 
\draw (5,0) -- (6,0) -- (5,0.5) -- cycle;
\draw [thick] (7,0) to [round right paren] (7,1);
\end{tikzpicture}}
\EndFor
\EndFor
\State{$Z \leftarrow \bm{\alpha}[0,l-1,R,C]$}
\end{algorithmic}
\end{algorithm*}

We then describe the \textbf{outside} algorithm to compute the outside tables in Algo. \ref{algo:outside}. In this algorithm, $\bigoplus$ stands for the \textit{logaddexp} operation.

\begin{algorithm*}[!htp]
\caption{Outside Algorithm}
\label{algo:outside}
\hspace*{\algorithmicindent} \textbf{Input: $\bm{S}, \bm{\alpha}$} \\
\hspace*{\algorithmicindent} \textbf{Output: $\bm{\beta}$} 
\begin{algorithmic}[1]
\State{$\bm{\beta}\leftarrow -\infty$}
\State{$\bm{\beta}[0,l-1,R,C] \leftarrow 0$} 
\For{$k \in l-1\dots 1$}
\For{$s \in 0\dots l-k$}
\State{$t=s+k$}
\For{$u \in s\dots t-1$}
\State{$\bm{\beta}[s, u+1, R, I]\leftarrow \bigoplus(\bm{\beta}[s, u+1, R, I], \bm{\beta}[s, t, R, C] + \bm{\alpha}[u+1, t, R, C])$}
\State{\begin{tikzpicture}[scale=0.5]
\draw [thick] (-1,0) to [round left paren] (-1,1);
\draw (1,0) -- (0,0) -- (0,1) -- cycle;
\node at (2,0.5) {$=$};
\draw[fill=gray!30] (4,0) -- (3,0) -- (3,1) -- (4,0.5) -- cycle;
\draw [-, thick] (4,0.5) -- (5,0.5); 
\draw (5,0) -- (6,0) -- (5,0.5) -- cycle;
\draw [thick] (7,0) to [round right paren] (7,1);
\end{tikzpicture}}
\EndFor
\For{$u \in s\dots t-1$}
\State{$\bm{\beta}[u+1, t, R, C]\leftarrow \bigoplus(\bm{\beta}[u+1, t, R, C], \bm{\beta}[s, t, R, C] + \bm{\alpha}[s, u+1, R, I])$}
\State{\begin{tikzpicture}[scale=0.5]
\draw [thick] (-1,0) to [round left paren] (-1,1);
\draw (1,0) -- (0,0) -- (0,1) -- cycle;
\node at (2,0.5) {$=$};
\draw (4,0) -- (3,0) -- (3,1) -- (4,0.5) -- cycle;
\draw [-, thick] (4,0.5) -- (5,0.5); 
\draw[fill=gray!30] (5,0) -- (6,0) -- (5,0.5) -- cycle;
\draw [thick] (7,0) to [round right paren] (7,1);
\end{tikzpicture}}
\EndFor
\If{$s>0$}
\For{$u \in s\dots t-1$}
\State{$\bm{\beta}[s, u, L, C]\leftarrow \bigoplus(\bm{\beta}[s, u, L, C], \bm{\beta}[s, t, L, C] + \bm{\alpha}[u, t, L, I])$}
\State{\begin{tikzpicture}[scale=0.5]
\draw [thick] (-1,0) to [round left paren] (-1,1);
\draw (1,0) -- (0,0) -- (1,1) -- cycle;
\node at (2,0.5) {$=$};
\draw[fill=gray!30] (4,0) -- (3,0) -- (4,0.5) -- cycle;
\draw [-, thick] (4,0.5) -- (5,0.5); 
\draw (5,0) -- (6,0) -- (6,1) -- (5,0.5) -- cycle;
\draw [thick] (7,0) to [round right paren] (7,1);
\end{tikzpicture}}
\EndFor
\For{$u \in s\dots t-1$}
\State{$\bm{\beta}[u, t, L, I]\leftarrow \bigoplus(\bm{\beta}[u, t, L, I], \bm{\beta}[s, t, L, C] + \bm{\alpha}[s, u, L, C])$}
\State{\begin{tikzpicture}[scale=0.5]
\draw [thick] (-1,0) to [round left paren] (-1,1);
\draw (1,0) -- (0,0) -- (1,1) -- cycle;
\node at (2,0.5) {$=$};
\draw (4,0) -- (3,0) -- (4,0.5) -- cycle;
\draw [-, thick] (4,0.5) -- (5,0.5); 
\draw[fill=gray!30] (5,0) -- (6,0) -- (6,1) -- (5,0.5) -- cycle;
\draw [thick] (7,0) to [round right paren] (7,1);
\end{tikzpicture}}
\EndFor
\EndIf
\For{$u \in s\dots t-1$}
\State{$\bm{\beta}[s, u, R, C] \leftarrow \bigoplus(\bm{\beta}[s, u, R, C], \bm{\beta}[s, t, R, I] + \bm{\alpha}[u+1, t, L, C] + \bm{S}[s, t])$}
\State{\begin{tikzpicture}[scale=0.5]
\draw [thick] (-1,0) to [round left paren] (-1,1);
\draw (1,0) -- (0,0) -- (0,1) -- (1,0.5) -- cycle;
\node at (2,0.5) {$=$};
\draw[fill=gray!30] (4,0) -- (3,0) -- (3,1) -- cycle;
\draw [->, thick] (3.25,1) to [out=30,in=150] (5.75,1); 
\draw (5,0) -- (6,0) -- (6,1) -- cycle;
\draw [thick] (7,0) to [round right paren] (7,1);
\end{tikzpicture}}
\EndFor
\For{$u \in s\dots t-1$}
\State{$\bm{\beta}[u+1, t, L, C]\leftarrow \bigoplus(\bm{\beta}[u+1, t, L, C], \bm{\beta}[s, t, R, I] + \bm{\alpha}[s, u, R, C] + \bm{S}[s, t])$}
\State{\begin{tikzpicture}[scale=0.5]
\draw [thick] (-1,0) to [round left paren] (-1,1);
\draw (1,0) -- (0,0) -- (0,1) -- (1,0.5) -- cycle;
\node at (2,0.5) {$=$};
\draw (4,0) -- (3,0) -- (3,1) -- cycle;
\draw [->, thick] (3.25,1) to [out=30,in=150] (5.75,1); 
\draw[fill=gray!30] (5,0) -- (6,0) -- (6,1) -- cycle;
\draw [thick] (7,0) to [round right paren] (7,1);
\end{tikzpicture}}
\EndFor
\If{$s>0$}
\For{$u \in s\dots t-1$}
\State{$\bm{\beta}[s, u, R, C] \leftarrow \bigoplus(\bm{\beta}[s, u, R, C], \bm{\beta}[s, t, L, I] + \bm{\alpha}[u+1, t, L, C] + \bm{S}[t, s])$}
\State{\begin{tikzpicture}[scale=0.5]
\draw [thick] (-1,0) to [round left paren] (-1,1);
\draw (1,0) -- (0,0) -- (0,0.5) -- (1,1) -- cycle;
\node at (2,0.5) {$=$};
\draw[fill=gray!30] (4,0) -- (3,0) -- (3,1) -- cycle;
\draw [<-, thick] (3.25,1) to [out=30,in=150] (5.75,1); 
\draw (5,0) -- (6,0) -- (6,1) -- cycle;
\draw [thick] (7,0) to [round right paren] (7,1);
\end{tikzpicture}}
\EndFor
\For{$u \in s\dots t-1$}
\State{$\bm{\beta}[u+1, t, L, C]\leftarrow \bigoplus(\bm{\beta}[u+1, t, L, C], \bm{\beta}[s, t, L, I] + \bm{\alpha}[s, u, R, C] + \bm{S}[t, s])$}
\State{\begin{tikzpicture}[scale=0.5]
\draw [thick] (-1,0) to [round left paren] (-1,1);
\draw (1,0) -- (0,0) -- (0,0.5) -- (1,1) -- cycle;
\node at (2,0.5) {$=$};
\draw (4,0) -- (3,0) -- (3,1) -- cycle;
\draw [<-, thick] (3.25,1) to [out=30,in=150] (5.75,1); 
\draw[fill=gray!30] (5,0) -- (6,0) -- (6,1) -- cycle;
\draw [thick] (7,0) to [round right paren] (7,1);
\end{tikzpicture}}
\EndFor
\EndIf
\EndFor
\EndFor
\end{algorithmic}
\end{algorithm*}

Finally, with the inside table $\bm{\alpha}$, outside table $\bm{\beta}$ and the marginalization $Z$ of all possible latent trees, we can compute the expectation of latent tree in an arc-decomposed manner. Algo. \ref{algo:marginal} describes the procedure. It results the matrix $\bm{P}$ containing the expectation of all individual arcs by marginalize over all other arcs except itself.

\begin{algorithm*}[!htp]
\caption{Arc Decomposed Expectation}
\label{algo:marginal}
\hspace*{\algorithmicindent} \textbf{Input: $\bm{\alpha},\bm{\beta}, Z$} \\
\hspace*{\algorithmicindent} \textbf{Output: $\bm{P}$}
\begin{algorithmic}[1]
\State{$\bm{P}\leftarrow 0$}
\For{$s \in 0\dots l-2$}
\For{$t \in s+1\dots l-1$}
\If{$s\neq t$}
\State{$\bm{P}[s, t]\leftarrow \exp(\bm{\alpha}[s,t,R,I]+\bm{\beta}[s,t,R,I]-Z)$}
\State{\begin{tikzpicture}
\draw [thick] (-1,0) to [round left paren] (-1,4);
\draw (0,0) -- (0,4) -- (8,0) -- (5.25, 0) -- (2.875,1.75) -- (2.875, 0) -- cycle;
\draw (3,0) -- (3,1.5) -- (5,0) -- cycle;
\node at (3.5,0.5) {$\alpha$};
\node at (1,2.5) {$\beta$};
\draw [thick] (9,0) to [round right paren] (9,4);
\end{tikzpicture}}
\If{$s>0$}
\State{$\bm{P}[t, s]\leftarrow \exp(\bm{\alpha}[s,t,L,I]+\bm{\beta}[s,t,L,I]-Z)$}
\State{\begin{tikzpicture}
\draw [thick] (-1,0) to [round left paren] (-1,4);
\draw (0,0) -- (0,4) -- (8,0) -- (4.125, 0) -- (4.125,1.75) -- (1.75, 0) -- cycle;
\draw (2,0) -- (4,0) -- (4,1.5) -- cycle;
\node at (3.5,0.5) {$\alpha$};
\node at (1,2.5) {$\beta$};
\draw [thick] (9,0) to [round right paren] (9,4);
\end{tikzpicture}}
\EndIf
\EndIf
\EndFor
\EndFor
\end{algorithmic}
\end{algorithm*}



\end{document}